\documentclass[10pt,twocolumn,letterpaper]{article}
\usepackage[]{cvpr} %
\usepackage[accsupp]{axessibility}  %

\definecolor{cvprblue}{rgb}{0.21,0.49,0.74}
\usepackage[pagebackref,breaklinks,colorlinks,allcolors=cvprblue]{hyperref}
\usepackage{array}
\def\paperID{14386} %
\def\confName{CVPR}
\def\confYear{2025}

\usepackage{newfloat}
\usepackage{listings}
\usepackage{etoolbox}
\DeclareCaptionStyle{ruled}{labelfont=normalfont,labelsep=colon,strut=off} %
\usepackage{amsmath}
\DeclareMathOperator*{\argmax}{arg\,max}

\usepackage{graphicx}
\usepackage{pgfplots}
\usepackage{subfiles} %
\usepackage{algorithm}
\usepackage[noend]{algpseudocode}

\usepackage{physics}
\usepackage{amsmath}
\usepackage{tikz}
\usepackage{mathdots}
\usepackage{yhmath}
\usepackage{cancel}
\usepackage{color}
\usepackage{siunitx}
\usepackage{array}
\usepackage{multirow}
\usepackage{amssymb}
\usepackage{gensymb}
\usepackage{tabularx}
\usepackage{extarrows}
\usepackage{booktabs}
\usetikzlibrary{fadings}
\usetikzlibrary{patterns}
\usetikzlibrary{shadows.blur}
\usetikzlibrary{shapes}

\DeclareMathOperator{\sgn}{sgn}

\DeclareMathSymbol{\mhyphen}{\mathord}{AMSa}{"39}

\usepackage{adjustbox}

\setcitestyle{numbers} %
\setcitestyle{numbers,open={[},close={]}} %
\title{Stop Walking in Circles! 
Bailing Out Early in Projected Gradient Descent 
}

\author{
    Philip Doldo \\  Booz Allen Hamilton \and Derek Everett \\  Booz Allen Hamilton \and Amol Khanna \\  Booz Allen Hamilton \and  Andre T Nguyen \\  Booz Allen Hamilton \and Edward Raff  \\  Booz Allen Hamilton
}

\begin{document}

\maketitle

\begin{abstract}
Projected Gradient Descent (PGD) under the $L_\infty$ ball has become one of the defacto methods used in adversarial robustness evaluation for computer vision (CV) due to its reliability and efficacy, making a strong and easy-to-implement iterative baseline. 
However, PGD is computationally demanding to apply, especially when using thousands of iterations is the current best-practice recommendation to generate an adversarial example for a single image. 
In this work, we introduce a simple novel method for early termination of PGD based on cycle detection by exploiting the geometry of how PGD is implemented in practice and show that it can produce large speedup factors while providing the \emph{exact} same estimate of model robustness as standard PGD. 
This method substantially speeds up PGD without sacrificing any attack strength,
enabling evaluations of robustness that were previously computationally intractable. 

\end{abstract}

\section{Introduction}

State-of-the-art computer vision classifiers have been shown to be vulnerable to 
\emph{adversarial examples}, 
small perturbations of their inputs which cause incorrect classifications~\cite{szegedy2013intriguing}. 
The existence of adversarial examples is particularly concerning in safety-critical applications and consequently, researchers have given significant attention to developing defenses against adversarial perturbations \cite{zhang2021adversarial, liang2022adversarial, akhtar2021advances, costa2024deep, chakraborty2021survey, silva2020opportunities}. %
As such, we are interested in reducing the time it takes for researchers and practitioners to perform standard evaluations of adversarial robustness, notably via the Projected Gradient Descent (PGD) attack under the $L_\infty$ ball. 

PGD is particularly demanding to evaluate in all cases because it is an iterative attack; for each sample $x$ we wish to perturb, $T$ forward and backward passes will be performed on top of the overhead of the optimization procedure. The use of $T \geq 1,000$ iterations has long been the recommendation \cite{athalye2018obfuscated,carlini2019evaluating} \footnote{See original recommendation from \url{https://nicholas.carlini.com/writing/2018/evaluating-adversarial-example-defenses.html}}. To put this computational burden in perspective, 100 training epochs on ImageNet is the same number of forward/backward passes as one test-set evaluation of ImageNet using $T=1{,}000$ PGD iterations. 

\begin{figure}[!h]
    \centering
    \adjustbox{max width=\columnwidth}{%
    \input{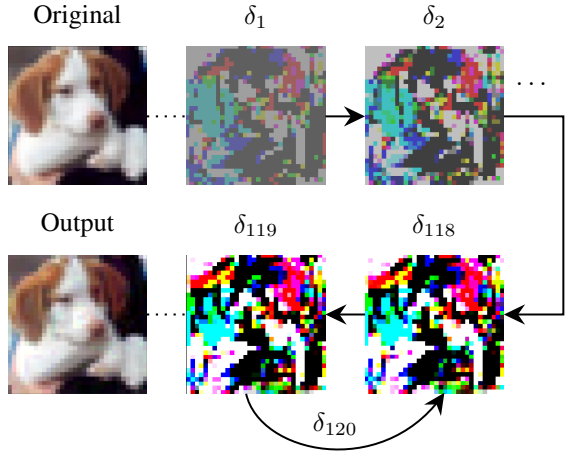}
    }
    \caption{The key insight of our work is that, under the constraint $\|\delta_t\|_\infty < \epsilon$, when it is not possible to find an adversarial example such that $f(x+\delta_t) \neq f(x)$, it is often the case that the PGD optimization will begin to cycle. That is for a given iterate $t$, and an offset $o$, we find $\delta_t = \delta_{t+o}$. We can save significant computational resources by detecting the cycles and stopping the PGD attack early. The perturbation $\delta$ in this figure is exaggerated for visibility.}
    \label{fig:pgd_cycles_example}
\end{figure}

Due to the prevalence of computer vision (CV) applications, the $L_\infty$ ball has become the predominant norm used in such evaluations. Inspired by this, we devise a simple strategy to reduce the number of PGD computations needed by up to 96\%, depending on the model being evaluated. Our key insight is that PGD is performed using a fixed step size $\alpha$. When an image $x$ is adversarially robust against a ball of radius $\epsilon$, this will often result in a cycling behavior as the iterates are pushed back to the feasible region, as shown by Figure \ref{fig:pgd_cycles_example}. By detecting these cycles, we can provably terminate the PGD attack, knowing that we will reach exactly the same conclusion of adversarial robustness. 

The rest of this manuscript is organized as follows. In Section \ref{sec:related_work} we will briefly review the related work to our own, and provide the technical background on PGD in Section \ref{sec:background}.
With this formalization, in Section \ref{sec:cycle-detection} we can state our new PGD with Cycle Detection, $\text{PGD}_{\text{CD}}$, that obtains identical results as the standard PGD attack at essentially no extra cost.
The worst-case scenario is that no cycle was detected, and so the same number of forward/backward passes are performed. We evaluate several models trained to be adversarially robust in Section \ref{ExperimentalResults} to demonstrate the range of results obtained with $\text{PGD}_{\text{CD}}$,
where we see that our method tends to perform best on larger and more expensive datasets like ImageNet compared to smaller datasets like CIFAR100.  In Section \ref{section:apgd} we will demonstrate how $\text{PGD}_{\text{CD}}$ can still achieve attack success rates comparable to, or even better, than adaptive step-size alternatives. This shows why PGD is still a strong and important baseline method, and why $\text{PGD}_{\text{CD}}$ should be the first attack used since it is both efficacious and orders of magnitude faster. Any sample quickly attacked by $\text{PGD}_{\text{CD}}$ need not be considered by slower algorithms, as the adversarial perturbation has already been found. 
Finally, we conclude in Section~\ref{sec:conclusion}.

\section{Related Work}\label{sec:related_work}

The scope of potential attack methods in use today is broad, and many attack algorithms are developed with different assumptions about attacker goals~\cite{NEURIPS2021_a709909b,MURA2025128918,Piras2023,CRECCHI2022257} and knowledge (white vs black box) and victim model type (e.g., deep learning vs decision trees). Despite the wide breadth, PGD remains a common component that is built upon due to its simplicity, and we will review how a number of key related works interface with PGD and its general use. 

Provably certifying a model's robustness generally requires solving expensive optimization problems~\cite{tjeng2017evaluating,Rahnama2020} or probabilistic and stochastic Monte Carlo bounds at inference time~\cite{cohen2019certified, salman2019provably,Raff_BaRT_2019}. In each case, they are computationally demanding for each prediction made by the model. Even when provable approaches are available, empirical evaluation is necessary to ensure that such implementations are (relatively) error-free and perform as expected~\cite{Fleshman2018a,NEURIPS2022_1add3bbd}.  
Adversarial training provides a state-of-the-art defense against adversarial perturbations \cite{ganin2016domain} and transfers all of the computational burdens to training time, imposing no extra cost on predictions. In such a case, the empirical evaluation is still demanding because $T$ iterations of forward/backward passes, when $T$ is often $ \geq 1{,}000$, is an exorbitant cost. This scenario makes empirical evaluation of the adversarial robustness all the more important, as there are no provable bounds to rely on.

Since the basic Fast Gradient Sign Method (FGSM)~\cite{goodfellow2014explaining} popularized adversarial attacks via white-box gradient-based search, improved iterative attacks have been a focus of study. Some, like seminal C\&W~\cite{7958570} attack purport better attack success given sufficient iterations, while others seek to produce smaller-norm examples in fewer iterations~\cite{NEURIPS2021_a709909b,Piras2023}. Still, the Projected Gradient Descent (PGD)~\cite{madry2017towards} has remained popular due to its efficacy, simplicity, and adaptability~\cite{Richards2021,everett2022}. 

Other iterative attacks that do not use the sign of the gradient
have been explored
\cite{Chen2022,kurakin2018adversarial}, but often found to be of comparable performance with minor tradeoffs once sufficient iterations are used~\cite{croce2020reliable,athalye2018obfuscated}. 

Also, 
the authors of 
Auto-PGD used an adaptive step size and momentum
and claimed that the resulting attack was stronger because it avoided cycles present in standard PGD~\cite{croce2020reliable}.
However, we demonstrate in Section~\ref{section:apgd} that Auto-PGD often performs similarly or even worse than PGD with a fixed step size.

Because PGD remains an effective and common baseline, even if other methods may perform better, we find it useful to improve the compute efficacy of PGD specifically as it has a desirable property of cycle formation to exploit, that is not shared by other methods. 

Marginally tighter estimates of a model's robust accuracy can be achieved by employing an ensemble of attacks, as is done by RobustBench \citet{croce2021robustbench}.
However, 
on ImageNet,
standard PGD with a fixed step size typically comes within 1\% of the robust accuracy obtained by RobustBench's ensemble, despite the ensemble being significantly more computationally expensive. 
We use RobustBench as a source of models trained to be adversarially robust
on which we perform evaluations,
and note that a faster PGD attack also accelerates these ensemble strategies.

\section{Background} \label{sec:background}

In this section, we will be mathematically more precise about the nature of adversarial attacks and PGD's operation. Let \(f : \mathcal{D} \to [0,1]^m\) be a model that takes an image \(X \in \mathcal{D}\) as input and maps it to a vector of probabilities \(f(X) \in [0,1]^m\) such that \(c(X) := \argmax_{i \in \{1, ..., m\}} f(X)\) is the %
predicted class label. 
An adversarial example of an image \(X\) with respect to a classifier \(c : \mathcal{D} \to \{1, ..., m\}\) is an image \(X_{\rm adv}\) which visually looks similar to \(X\), yet gets misclassified by the model. That is, \(X_{\rm adv}\ = X + \delta\) where \(\|\delta\| \leq \epsilon\) and \(c(X) \neq c(X_{\rm adv})\) for some norm \(\|\cdot\|\) and radius \(\epsilon > 0\). We refer to $\delta$ as an \emph{adversarial perturbation}, 
and typically the norm of the perturbation should be sufficiently small such that it is difficult for a human observer to distinguish between the original and adversarial images.
We note that we only consider the concept of an adversarial example to be meaningful if the unperturbed image is correctly classified by the model to begin with.

In this work, we focus on 
\emph{untargeted} evasion attacks,
where the goal of the adversary is to perturb an image such that it gets classified as any incorrect class label. 
Additionally, we consider an \(L_{\infty}\) adversarial threat model, 
which requires
that \(\|\delta\|_{\infty} \leq \epsilon\) so that the value of each pixel is within an \(\epsilon\) radius of the original pixel value. 
We note that for an image classifier to achieve a high classification accuracy in practice, it is typical that it expects images that are preprocessed and scaled appropriately and so when we talk about being within an \(\epsilon\) radius of the original image we are assuming that the original image was already preprocessed and scaled accordingly.

\subsection{Evaluating Robustness}

In practice, we can only empirically evaluate the robustness of a model with respect to a particular finite dataset of images \( \mathcal{D}_{\rm data} := \{X_1, ..., X_{N_{\rm data}}  \}\). 
The \emph{clean accuracy} of a classifier with respect to a dataset is the percentage of images in that dataset that are correctly classified by the model. 
The \emph{robust accuracy} of a classifier \(c : \mathcal{D} \to \{1, ..., m\}\) with respect to a dataset \(\mathcal{D}_{\rm data}\) and an \(\epsilon\)-radius \(L_{\infty}\) constraint is the percentage of images \(X \in \mathcal{D}_{\rm data}\) such that \(c(X) = c(X+\delta)\) for all \(\|\delta\|_{\infty} \leq \epsilon\) and \(X\) is correctly classified in the clean setting.

In general, verifying that an image retains its predicted class label under any perturbation in the \(L_{\infty}\) ball is a computationally intractable problem, hence we must resort to obtaining an upper bound on the model's true robust accuracy on a given dataset. 
To obtain such an upper bound, we must perform an attack -- a concrete optimization or search algorithm -- 
to generate perturbations for each (correctly classified) image in the dataset.
Stronger attacks cause more misclassifications and
thus result in a lower reported robust accuracy and tighter upper bound. 

There are typically tradeoffs between an attack's strength and its computational cost, 
and so it is desirable to develop computationally efficient attacks that yield reasonably tight upper bounds that can be used as a baseline for evaluating robustness. 

\subsection{Projected Gradient Descent}

Suppose we have a differentiable loss function \(\mathcal{L}\) 
so that \(\mathcal{L}(f(X), y)\) is the loss of the model \(f\) for the image \(X\) and its corresponding true class label \(y\). 
In the white-box adversarial attacks we consider, an adversary performs an optimization over \(\delta\) such that \(\mathcal{L}(f(X+\delta), y)\) is maximized subject to a constraint on the norm of \(\delta\). 
When the loss function is the cross entropy loss between the true and predicted label distributions, this is maximized by predicting any incorrect label with maximal confidence. 

We focus our attention on the following constrained maximization problem: 
\begin{equation}
\max_{\delta} \mathcal{L}(f(X+\delta), y) \hspace{2mm} \text{such that} \hspace{2mm} \|\delta\|_{\infty} \leq \epsilon.
\end{equation} 
A strong gradient-based approach under an \(L_{\infty}\) constraint on \(\delta\) is Projected Gradient Descent (PGD), 
where the perturbation \(\delta\) is constructed by iterative steps in the direction of the \emph{sign} of the gradient, %
and any step taken outside the \(L_\infty\) ball is always projected back to its surface. 
PGD uses the following update rule: 
\begin{equation}
\delta^{(i+1)} = \mathcal{P}_{\mathcal{B}}
\left( 
    \delta^{(i)} + \alpha \hspace{1mm} \text{sign}
        \left(
        \nabla_X( \mathcal{L}(f(X+\delta^{(i)}), y ) \right) 
    \right),
\label{PGD:update}
\end{equation}
where \(\mathcal{P}_{\mathcal{B}}\) projects its input to the nearest point on the ball \(\mathcal{B} = \{\delta : \|\delta\|_\infty \leq \epsilon\}\) and \(\alpha > 0\) is the step size. In the simplest form of PGD, we initialize the perturbation \(\delta^{(0)}\) at the origin and use a constant step size \(\alpha\). 

Returning to the tradeoff between attack cost and strength, strong PGD attacks may require a ceiling of
thousands of iterations before
the adversarial robustness is accurately estimated -- a large computational expense.
In what follows, we describe how we exploit the behavior of PGD's fixed step size 
to significantly accelerate computation in many situations without sacrificing \emph{any} attack strength, 
yielding a much more efficient baseline for evaluating model robustness. 
Efficiently computing such a baseline is important because there is a tradeoff between attack strength and required compute resources. A stronger attack may be undesirable if it is too expensive and therefore it is important that the most efficient implementation of a baseline attack is used in order to get a fair comparison.

\section{Our Method for Cycle Detection and Termination} \label{sec:cycle-detection}
Despite its simplicity, PGD with a fixed step size is an effective attack in many situations and is straightforward to implement, making it an important baseline for comparing other attacks. 
However, running PGD for \(T\) iterations on every image in a dataset can be prohibitively expensive for reasonably large \(T\). 
Our faster strategy relies on two simple insights that, to wit,  have not been previously documented in the literature. 

1. When an input $\boldsymbol{x}$ is robust under the $L_\infty$ ball of radius $\epsilon$,  it is often the case that cycles will form. In short, the mass in high dimensional spaces is majority on the edge of the ball, and the fixed step size of PGD creates a discrete set of options. 

2. When an input $\boldsymbol{x}$ can be attacked successfully, it often requires less than 10 iterations to find the $\delta$. Standard PGD implementations of 
Foolbox\footnote{\url{https://tinyurl.com/foolboxPGD}},
cleverhans\footnote{\url{https://tinyurl.com/cleverhansPGD}},
and ART\footnote{\url{https://tinyurl.com/advrtPGD}}
do not check that an adversarial example has been found during the iteration, and so all $T$ iterations will be performed. 

We have experimentally observed that the number of required iterations to trick an undefended ResNet50 ImageNet model on a given image is typically quite small, with a median of only three iterations using a step size of \(\epsilon/4\) with \(\epsilon=4/255\). 
Consequently, the robust accuracy of undefended models can often be driven very close to zero with little required compute. 
Additionally, even for defended models, it turns out that most of the images that 
are misclassified by PGD require a relatively small number of iterations compared to the ceiling. We will detail these further in our experimental results in Section \ref{ExperimentalResults}.

That being said, some images can require thousands of iterations before being misclassified. 
This becomes a computational burden when evaluating the robustness of defended models; 
strong defenses will prevent the misclassification of a large percentage of the dataset, 
and we have no way of knowing a priori whether the attack will cause a misclassification on any given image. 
Therefore, it would seem that to obtain a tight estimate of robust accuracy, 
we have no recourse but to use the entire iteration budget on all the images 
that will not be misclassified. 
However, the following insights into the iterative PGD method 
will allow us to identify whether PGD has failed to cause misclassification much before the iteration ceiling is reached. 
This allows us to abort attacking an image 
on which further PGD iterations will never produce a misclassification, saving
a significant amount of computation time without sacrificing any attack strength. 
Indeed, we have experimentally observed that the number of iterations required by our attack is often smaller than that required by standard PGD by a factor of 10 on multiple ImageNet models, which results in significant computational savings. We describe our method and its motivation in the following subsection.

\subsection{Early Termination via Cycle Detection}

Our approach is straightforward to implement: while traversing the confines of the \(L_\infty\) ball via PGD with a fixed step size,
terminate PGD iterations on an image if you ever revisit a point within the \(L_\infty\) ball.
See Algorithm \ref{PGDalg} for a step-by-step description. We note that in practice, we do not actually store all of the encountered perturbations to check for a cycle, but instead, we compute and store a hash of each perturbation tensor, computed with the Python syntax \texttt{hash(tuple(x.flatten().tolist()))} for a PyTorch tensor \texttt{x}, for more efficient memory management and look-ups. 
Essentially, if you ever revisit a point in the image space,  
then you are guaranteed to follow the signed gradient in the same sequence of iterations you did the last time you visited it, 
and are, therefore, stuck in a closed cycle that can never cause a misclassification. 

This is because PGD uses a fixed step size and only the local gradient information
and does not employ methods such as momentum, which could 
prevent
these cycles. 
Per our related work, we rely on the fact that PGD with fixed step sizes is competitive with adaptive learning rates for finding adversarial examples in terms of both iterations and efficacy.

\begin{algorithm}[!h]
    \caption{PGD with Cycle Detection}
    \label{PGDalg}
\begin{algorithmic}[1]
    \State \textbf{Input:} model \(f\), image \(X\), label \(y\), radius \(\epsilon\), iteration budget \(T_{\rm iter}\), step size \(\alpha\), loss \(\mathcal{L}\)
    \State \textbf{Output:} return True if the model is tricked and return False otherwise
    \State \textbf{Initialize} \(\mathcal{S} \gets \emptyset\)\Comment{Set of observed perturbations}
    \State \textbf{Initialize} \(\delta^{(0)} \gets \mathbf{0}\)\Comment{Perturbation starts as zero tensor} 
    \For{\(i=1, ..., T_{\rm iter}\)}
        \State $d \gets \nabla_X( \mathcal{L}(f(X+\delta^{(i-1)}), y )$
        \State \(\delta^{(i)} \gets \mathcal{P}_{\mathcal{B}}
    \left( 
    \delta^{(i-1)} + \alpha \hspace{1mm} \text{sign}
        \left(
        d \right) 
    \right)\) \Comment{PGD Update}
    \If{\(c(X+\delta^{(i)}) \neq y\)}
        \State \textbf{return} True \Comment{The model is tricked}
    \ElsIf{\(\delta^{(i)} \in \mathcal{S}\)}
        \State \textbf{return} False \Comment{A cycle is encountered}
    \Else
        \State \(\mathcal{S} \gets \mathcal{S} \cup \{ \delta^{(i)} \}\) \Comment{Store the perturbation}
    \EndIf
    \EndFor
    \State \textbf{return} False \Comment{Return False if the model is not tricked within \(T_{\rm iter}\) iterations}
\end{algorithmic}
\end{algorithm}
\begin{figure}[!h]
\adjustbox{max width=\columnwidth}{%
\tikzset{every picture/.style={line width=0.75pt}} %

\begin{tikzpicture}[x=0.75pt,y=0.75pt,yscale=-1,xscale=1]
\draw  [color={rgb, 255:red, 0; green, 0; blue, 0 }  ,draw opacity=1 ][fill={rgb, 255:red, 0; green, 0; blue, 0 }  ,fill opacity=1 ] (95,225) .. controls (95,219.48) and (99.48,215) .. (105,215) .. controls (110.52,215) and (115,219.48) .. (115,225) .. controls (115,230.52) and (110.52,235) .. (105,235) .. controls (99.48,235) and (95,230.52) .. (95,225) -- cycle ;
\draw  [color={rgb, 255:red, 74; green, 74; blue, 74 }  ,draw opacity=1 ][fill={rgb, 255:red, 74; green, 74; blue, 74 }  ,fill opacity=1 ] (40,130) .. controls (40,124.48) and (44.48,120) .. (50,120) .. controls (55.52,120) and (60,124.48) .. (60,130) .. controls (60,135.52) and (55.52,140) .. (50,140) .. controls (44.48,140) and (40,135.52) .. (40,130) -- cycle ;
\draw  [color={rgb, 255:red, 74; green, 74; blue, 74 }  ,draw opacity=1 ][fill={rgb, 255:red, 74; green, 74; blue, 74 }  ,fill opacity=1 ] (140,130) .. controls (140,124.48) and (144.48,120) .. (150,120) .. controls (155.52,120) and (160,124.48) .. (160,130) .. controls (160,135.52) and (155.52,140) .. (150,140) .. controls (144.48,140) and (140,135.52) .. (140,130) -- cycle ;
\draw  [color={rgb, 255:red, 0; green, 0; blue, 0 }  ,draw opacity=1 ][fill={rgb, 255:red, 0; green, 0; blue, 0 }  ,fill opacity=1 ] (90,29.5) .. controls (90,23.98) and (94.48,19.5) .. (100,19.5) .. controls (105.52,19.5) and (110,23.98) .. (110,29.5) .. controls (110,35.02) and (105.52,39.5) .. (100,39.5) .. controls (94.48,39.5) and (90,35.02) .. (90,29.5) -- cycle ;
\draw [color={rgb, 255:red, 144; green, 19; blue, 254 }  ,draw opacity=1 ][line width=2.25]    (150,130) -- (53.9,53.12) ;
\draw [shift={(50,50)}, rotate = 38.66] [fill={rgb, 255:red, 144; green, 19; blue, 254 }  ,fill opacity=1 ][line width=0.08]  [draw opacity=0] (8.57,-4.12) -- (0,0) -- (8.57,4.12) -- cycle    ;
\draw [color={rgb, 255:red, 74; green, 144; blue, 226 }  ,draw opacity=1 ][line width=2.25]    (50,50) -- (50,125) ;
\draw [shift={(50,130)}, rotate = 270] [fill={rgb, 255:red, 74; green, 144; blue, 226 }  ,fill opacity=1 ][line width=0.08]  [draw opacity=0] (8.57,-4.12) -- (0,0) -- (8.57,4.12) -- cycle    ;
\draw [color={rgb, 255:red, 208; green, 2; blue, 27 }  ,draw opacity=1 ][line width=2.25]    (50,130) -- (86.88,83.9) ;
\draw [shift={(90,80)}, rotate = 128.66] [fill={rgb, 255:red, 208; green, 2; blue, 27 }  ,fill opacity=1 ][line width=0.08]  [draw opacity=0] (8.57,-4.12) -- (0,0) -- (8.57,4.12) -- cycle    ;
\draw [color={rgb, 255:red, 144; green, 19; blue, 254 }  ,draw opacity=1 ][line width=2.25]    (50,130) -- (146.1,53.12) ;
\draw [shift={(150,50)}, rotate = 141.34] [fill={rgb, 255:red, 144; green, 19; blue, 254 }  ,fill opacity=1 ][line width=0.08]  [draw opacity=0] (8.57,-4.12) -- (0,0) -- (8.57,4.12) -- cycle    ;
\draw [color={rgb, 255:red, 74; green, 144; blue, 226 }  ,draw opacity=1 ][line width=2.25]    (150,50) -- (150,125) ;
\draw [shift={(150,130)}, rotate = 270] [fill={rgb, 255:red, 74; green, 144; blue, 226 }  ,fill opacity=1 ][line width=0.08]  [draw opacity=0] (8.57,-4.12) -- (0,0) -- (8.57,4.12) -- cycle    ;
\draw [color={rgb, 255:red, 208; green, 2; blue, 27 }  ,draw opacity=1 ][line width=2.25]    (150,130) -- (113.12,83.9) ;
\draw [shift={(110,80)}, rotate = 51.34] [fill={rgb, 255:red, 208; green, 2; blue, 27 }  ,fill opacity=1 ][line width=0.08]  [draw opacity=0] (8.57,-4.12) -- (0,0) -- (8.57,4.12) -- cycle    ;
\draw    (10,230) -- (10,130) -- (200,130) -- (200,230) ;

\draw (215,200) node  [rotate=-90] [align=left] {\begin{minipage}[lt]{95.2pt}\setlength\topsep{0pt}
$\displaystyle L_{\infty }$ ball boundary
\end{minipage}};
\draw (255,25) node   [align=left] {\begin{minipage}[lt]{101.83pt}\setlength\topsep{0pt}
The solution $\displaystyle \left( a^{*} ,b^{*}\right)$ is outside the $\displaystyle \epsilon $ ball.
\end{minipage}};
\draw (300,160) node   [align=left] {\begin{minipage}[lt]{95.2pt}\setlength\topsep{0pt}
\textcolor[rgb]{0.82,0.01,0.11}{1. Compute direction $\displaystyle \nabla f\left(\tilde{\boldsymbol{x}}\right)$}\\\textcolor[rgb]{0.56,0.07,1}{2. Direction becomes $\displaystyle \alpha \ \sgn\left( \nabla f\left(\tilde{x}\right)\right)$}\\\textcolor[rgb]{0.29,0.56,0.89}{3. Projection $\displaystyle \mathcal{P}$ back to feasible space}
\end{minipage}};
\draw (50.81,104.52) node [anchor=north west][inner sep=0.75pt]  [font=\tiny,rotate=-307.32]  {$\textcolor[rgb]{0.82,0.01,0.11}{\nabla f\left(\boldsymbol{\tilde{x}}_{1}\right)}$};
\draw (131.71,82.32) node [anchor=north west][inner sep=0.75pt]  [font=\tiny,rotate=-49.22]  {$\textcolor[rgb]{0.82,0.01,0.11}{\nabla f}\textcolor[rgb]{0.82,0.01,0.11}{(}\textcolor[rgb]{0.82,0.01,0.11}{\tilde{\boldsymbol{x}}_{2}}\textcolor[rgb]{0.82,0.01,0.11}{)}$};
\draw (92.3,76.55) node [anchor=north west][inner sep=0.75pt]  [font=\tiny,rotate=-320.5]  {$\textcolor[rgb]{0.56,0.07,1}{\alpha \sgn\left( \nabla f\left(\boldsymbol{\tilde{x}}_{1}\right)\right)}$};
\draw (151,104.4) node [anchor=north west][inner sep=0.75pt]  [font=\footnotesize,color={rgb, 255:red, 74; green, 144; blue, 226 }  ,opacity=1 ]  {$\delta _{1} =\mathcal{P}( \cdot )$};
\draw (118,142.4) node [anchor=north west][inner sep=0.75pt]    {$\tilde{\boldsymbol{x}}_{2} =( a_{2} ,\epsilon )$};
\draw (11,142.4) node [anchor=north west][inner sep=0.75pt]    {$\tilde{\boldsymbol{x}}_{1} =( a_{1} ,\epsilon )$};
\draw (105,225) node  [color={rgb, 255:red, 255; green, 255; blue, 255 }  ,opacity=1 ]  {$x$};
\draw (100,29.5) node  [color={rgb, 255:red, 255; green, 255; blue, 255 }  ,opacity=1 ]  {$x^{*}$};

\end{tikzpicture}
}
\caption{
A 2D example of how a cycle of length two can occur on the boundary of the $L_\infty$ ball. The minimum-norm adversarial example $\boldsymbol{x}^*$ is outside the $\epsilon$ ball, so the original datum $\boldsymbol{x}$ is robust. The gradients point toward the optimal solution so that the projected signed gradient steps oscillate between the same two points on the boundary, one on each side of the optimum. %
}
\label{LinfBall}
\end{figure}
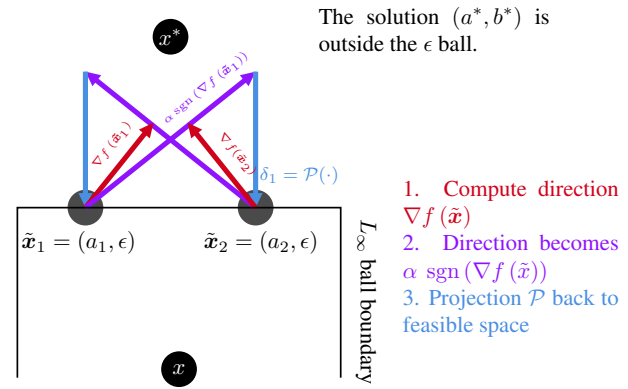
Our cycle detection and termination method works well in practice, and this may be due to the 
behavior
of the PGD algorithm described below. 
At each iteration, the gradient points in the local direction of steepest ascent of the loss function, and PGD takes the sign of the gradient. 
If following the vector given by the step size times the signed gradient results in a point outside of the \(L_\infty\) ball, then the point will get projected back to the boundary of the \(L_\infty\) ball. 
If this occurs on a flat surface or edge of the \(L_\infty\) ball, then PGD can often be caught in a cycle of length two, perhaps perpetually chasing a local maximum outside of the \(L_\infty\) ball while being stuck on its boundary.
We provide a two-dimensional visualization of this phenomenon in Figure \ref{LinfBall}.

Empirically we observed this failure mode to happen frequently, sometimes seeing even more complicated and longer cycles on the boundary of the \(L_\infty\) ball. 
Interestingly, we observed that cycles of length four or longer appeared much more commonly in the low-dimensional CIFAR10 and CIFAR100 datasets than in the higher-dimensional ImageNet dataset. 

In theory, it is also possible that a local extremum of the loss function exists within the interior of the \(L_\infty\) ball, and in this case, PGD, and signed gradient ascent with a fixed step size more broadly, can continue to iterate without converging. 
In this case, while decreasing the step size would allow signed gradient descent to converge to the \emph{local} maximum~\cite{moulay2019properties}, this point might only cause lower confidence and not misclassification. 
However, we did not observe cycles in the interior of the \(L_\infty\) ball and we suspect that this is because most of the volume of the \(L_\infty\) ball is concentrated toward its boundary in high dimensions.

\begin{figure}[!h]
  \includegraphics[width=.95\linewidth, height=.17\paperheight]{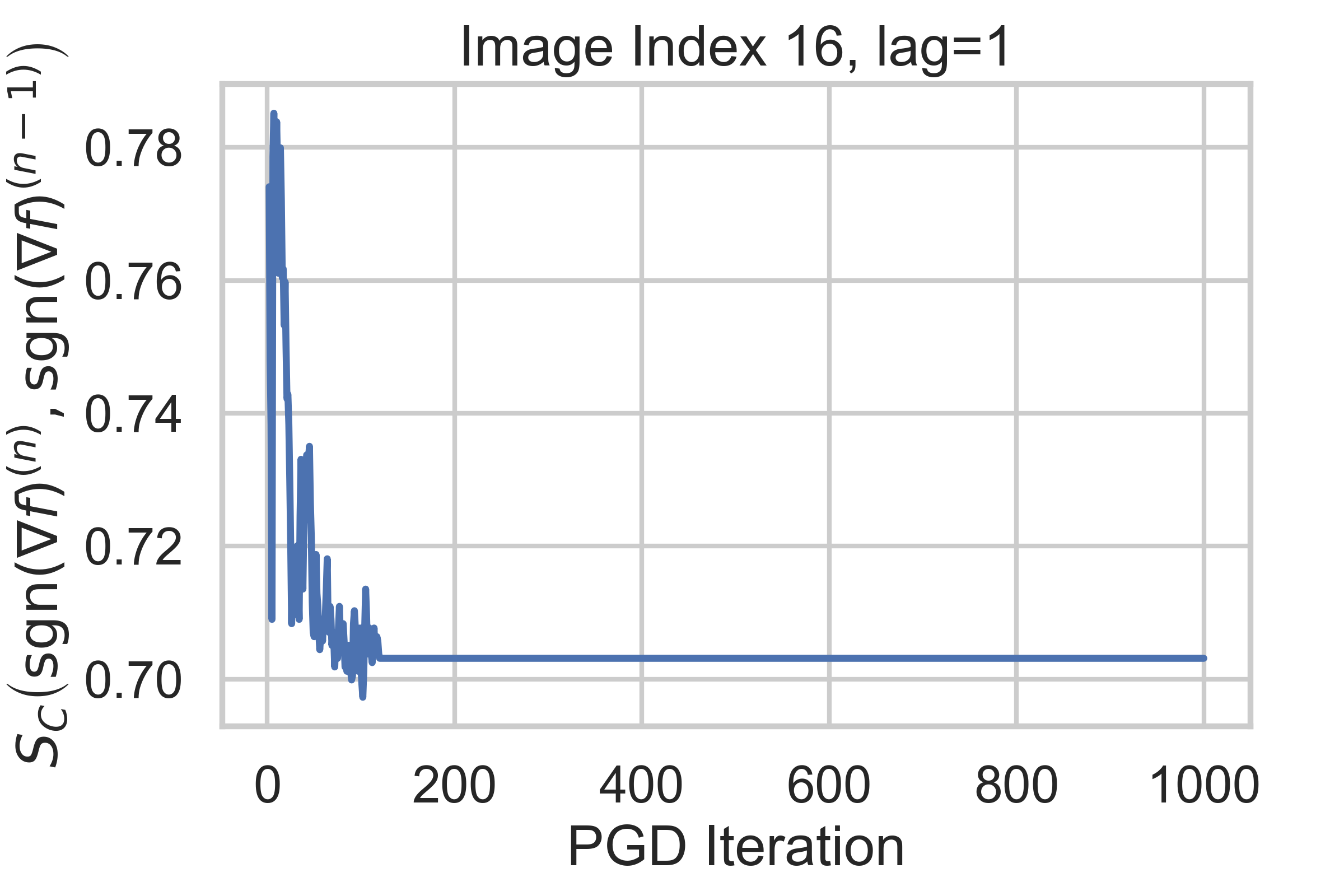}
\\
  \includegraphics[width=.95\linewidth, height=.17\paperheight]{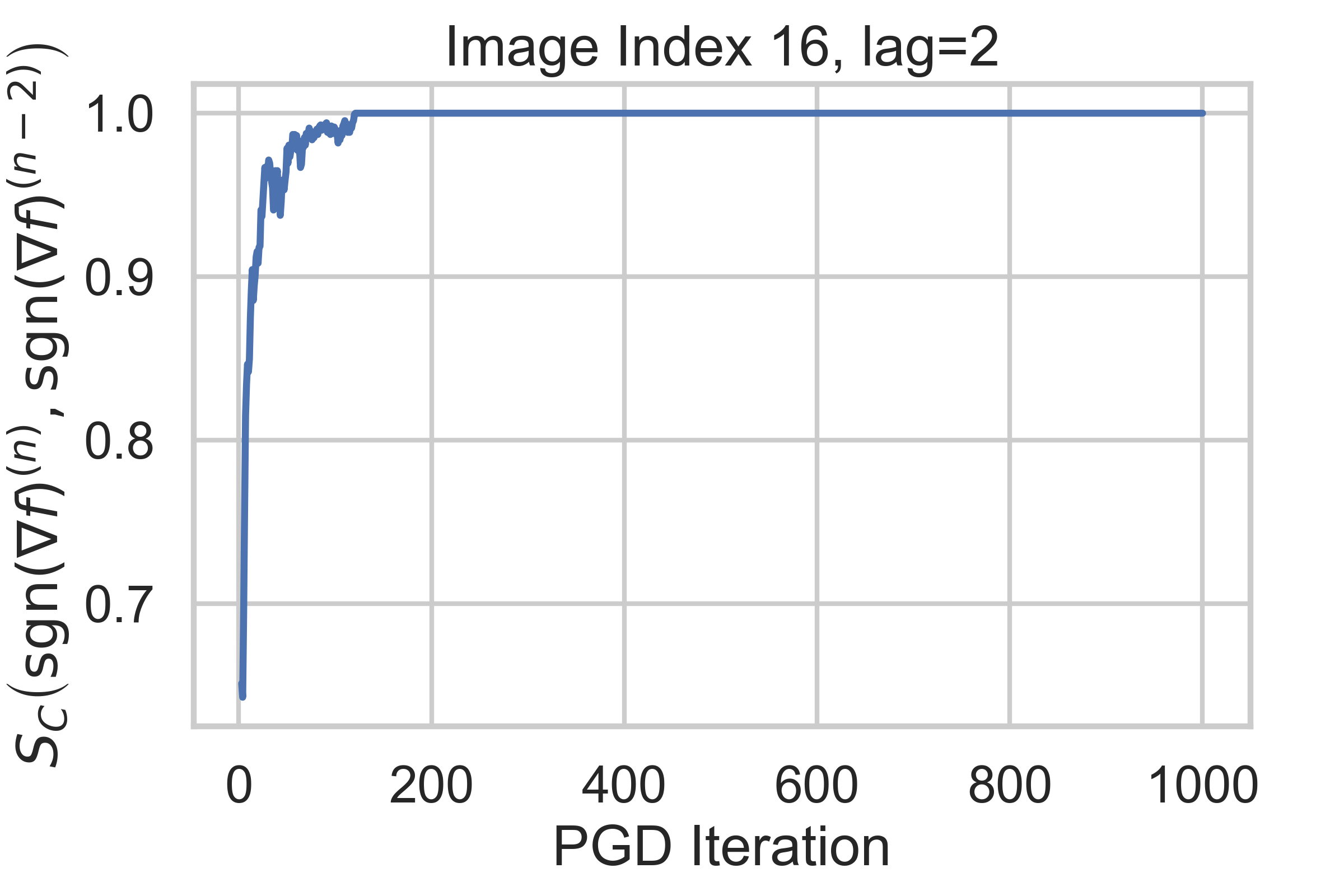}
\caption{Cosine similarities ($S_C$) between signed gradients using the \emph{Carmon2019Unlabeled} \citet{carmon2019unlabeled} CIFAR10 model from RobustBench. The $S_C$ between the 
first and second preceding items (lag=1, 2)
all converge to a constant when a cycle forms. When the lag is a multiple of the cycle length, we will, by definition, reach a $S_C$ of 1 as it compares to previous iterations of itself.
}
\label{CosineSimilarities}
\end{figure}

Cycles can also be observed in the PGD search by identifying when the  cosine similarity,
\(S_C(x,y) := \frac{x^\top y}{\|x\|_2 \|y\|_2} \),
of successive signed gradients converges to a constant value. 
As an example, in Figure \ref{CosineSimilarities} we show a cycle of length two that was encountered when running PGD on a CIFAR10 model. 
After sufficient PGD iterations occur and PGD gets stuck in a cycle, the cosine similarity of successive signed gradients converges to a value close to 0.7, indicating that the successive signed gradients are different from each other, but the same two signed gradients are always compared after enough PGD iterations have passed. Similarly, the cosine similarity between every other signed gradient converges to one, indicating that every other signed gradient is identical to one another after enough PGD iterations. %

Although a cosine similarity of unity implies only that the signed gradients are equivalent up to a scale factor, we investigated the underlying tensors in these cases and found them to in fact be identical as all entries of the gradients were nonzero up to floating point precision so that the signed gradients all had the same magnitude.
This observation and convergence of $S_C$ may be used for inexact early-cycle breaking, but in this work, we constrain ourselves to exact PGD equivalence.

\begin{figure}[!h]
\centering
    \includegraphics[width=.95\linewidth]{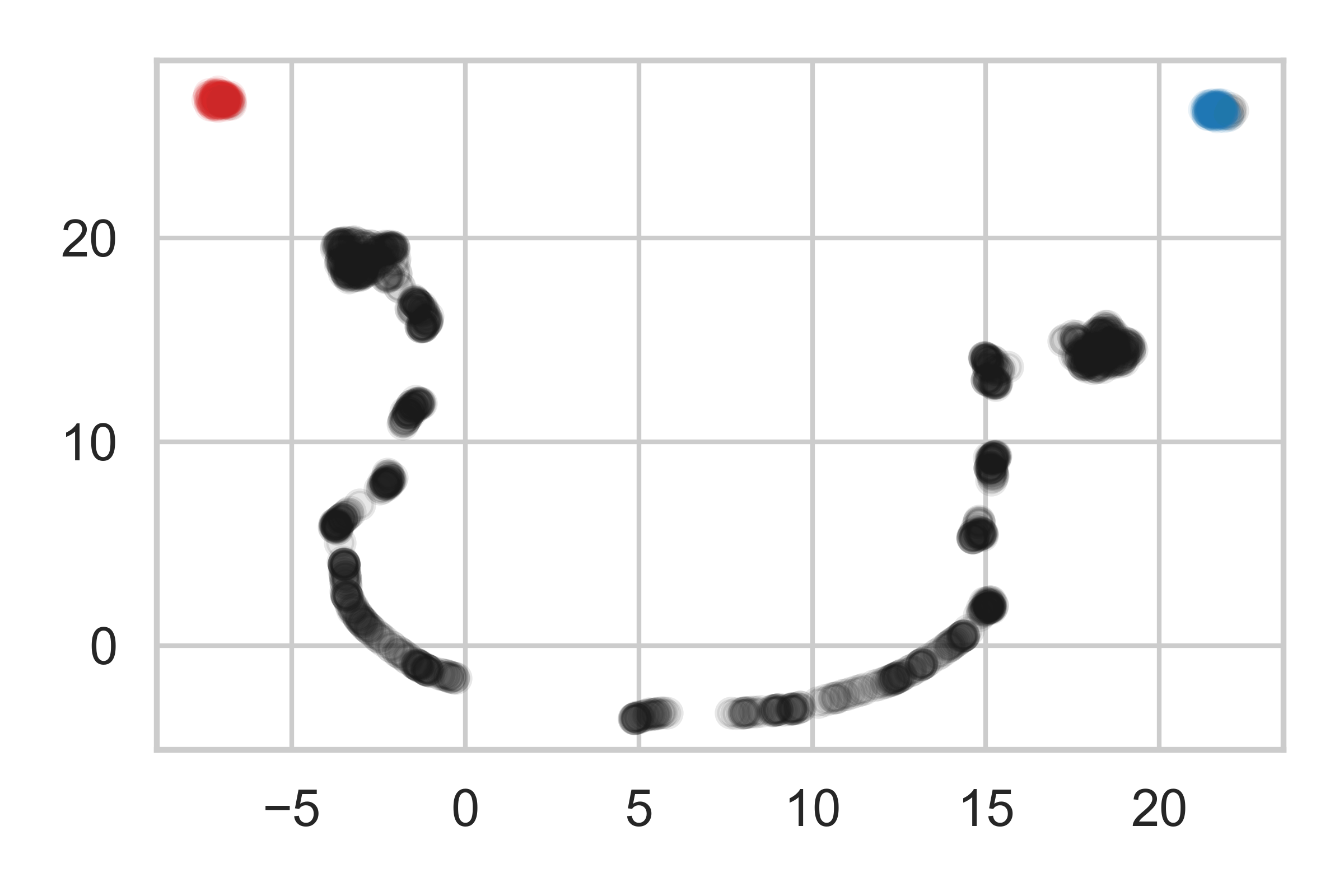}
\\
    \includegraphics[width=.95\linewidth]{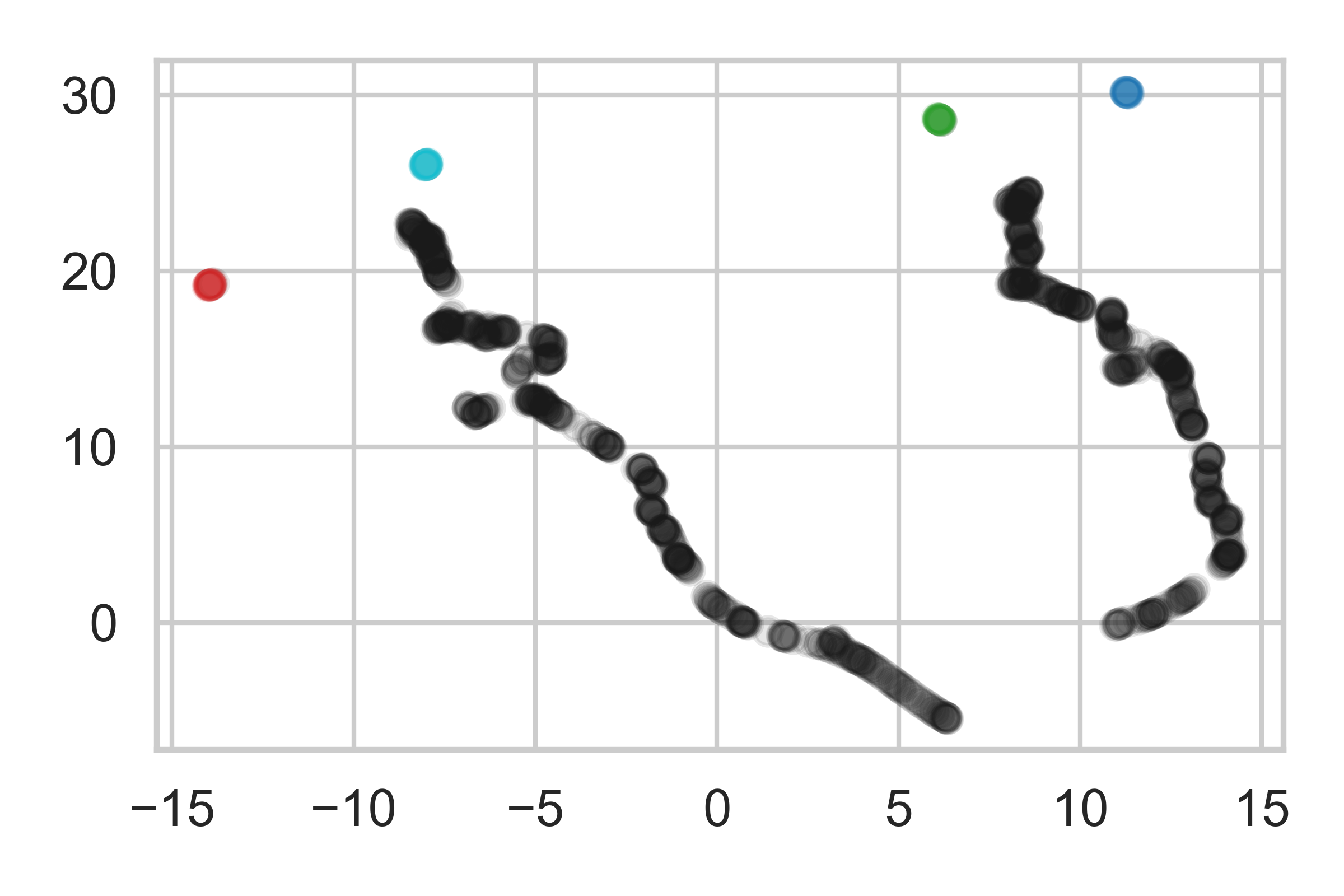}
\caption{Two-dimensional UMAP projection of perturbations \(\delta^{(i)}\) generated from different PGD iterations for the \emph{Carmon2019Unlabeled} CIFAR10 model from RobustBench. \textbf{Top}: 1000 PGD iterations are shown, with a length-two cycle occurring after 845 iterations and alternating between the red and blue points. 
\textbf{Bottom}: 1000 PGD iterations are shown, and a length-four cycle occurs after 934 iterations and cycles across the green, cyan, blue, and red points (in that order). 
Black points correspond to perturbations before the cycle begins, and colored points correspond to the perturbations between which PGD ultimately oscillates.}
\label{umap}
\end{figure}

\subsubsection*{Hashing on the CPU} Using the naive hashing method as described earlier is slow due to moving data from GPU to CPU. 
A custom hash function written in CUDA would run fastest, but is not easily available in PyTorch. For simplicity, we use a fast-enough function by computing, essentially, \texttt{hash(torch.frexp($\boldsymbol{h}^\top \delta^{(i)}$))}, where $\boldsymbol{h}$ is a vector of normally distributed values from $\mathcal{N}(0, 1/\sqrt{d})$. The use of \texttt{frexp} avoids denormalization issues that may be non-deterministic on some machines and results in only a small amount of data going to the CPU to be hashed (in reality, we hash the tuple of floats obtained by calling \texttt{.item()} on both the mantissa and exponent tensors outputted from \texttt{torch.frexp}, which moves these floats to the CPU).%

\section{Experimental Results}
\label{ExperimentalResults}

We experimentally demonstrate the success of our approach by attacking defended models from RobustBench  \cite{croce2021robustbench} on the CIFAR10, CIFAR100, and ImageNet datasets, providing independently developed, validated, and top-performing robust models. We run PGD with and without cycle detection and compare the number of iterations required to attack every image in the test dataset. Recall that each iteration requires a gradient computation and thus performs both a forward and backward pass on the model being evaluated; this makes up the majority of the computational burden of PGD. 
We will show that our method, \(\rm PGD_{\rm CD}\), obtains the exact same result as PGD but in far fewer iterations for most models.

Following recommended practices, we use a maximum of \(T_{\rm iter}=1000\) iterations on each image~\cite{athalye2018obfuscated, carlini2019evaluating}. We use an attack radius of \(\epsilon=8/255\) for CIFAR10 and CIFAR100 models and \(\epsilon=4/255\) for ImageNet models as these are the respective values that RobustBench claims robustness for. We use a step size of \(\epsilon/4\) because it gave relatively good results and others have demonstrated that it is often a good choice, see e.g. Table 9 in \cite{croce2020reliable}, though it is worth noting that this is not the optimal step size for all models used in these experiments. 

\begin{figure}[!h]
\adjustbox{max width=0.49\columnwidth}{%
  \includegraphics[]{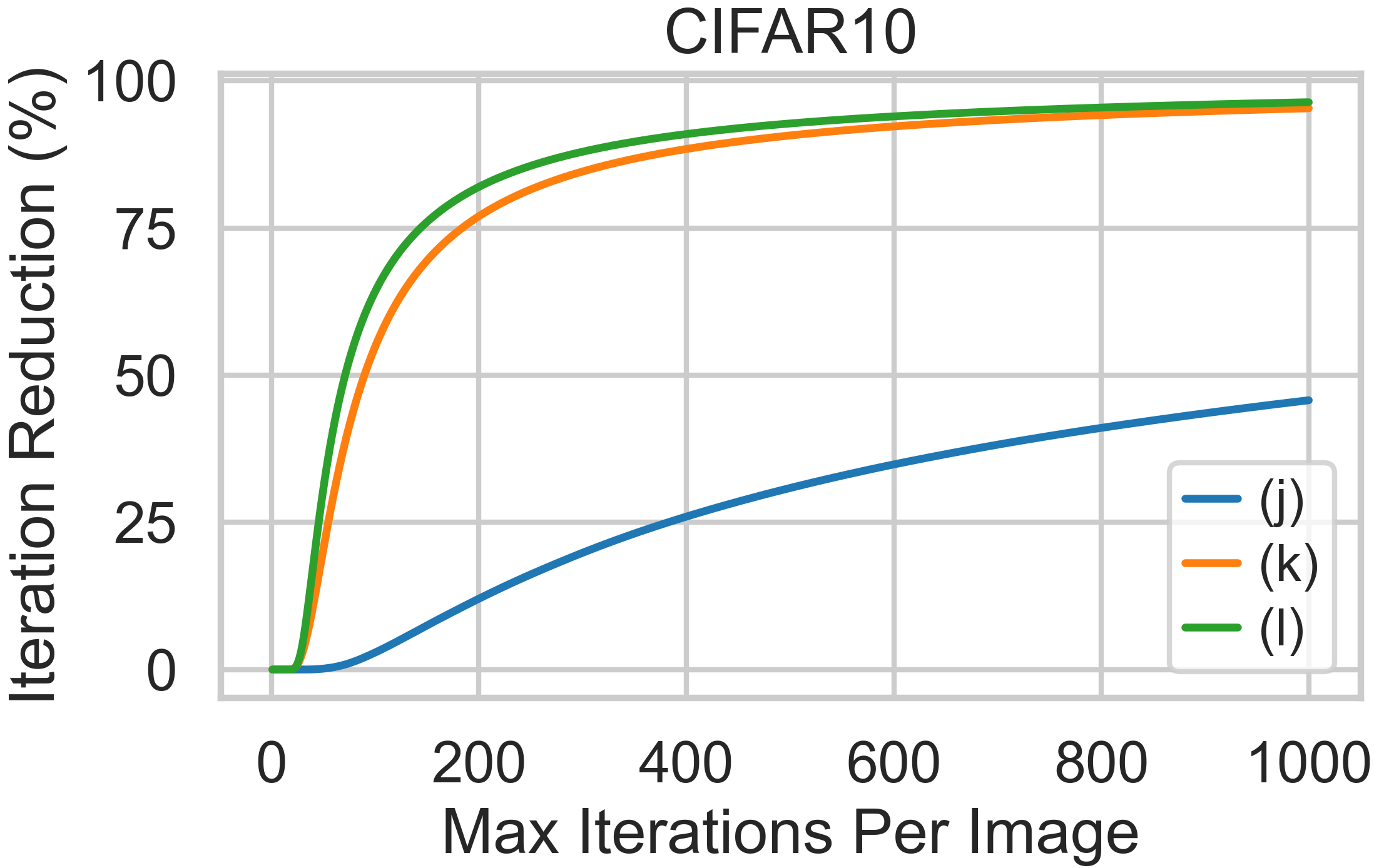}
  }
\adjustbox{max width=0.49\columnwidth}{%
  \includegraphics[]{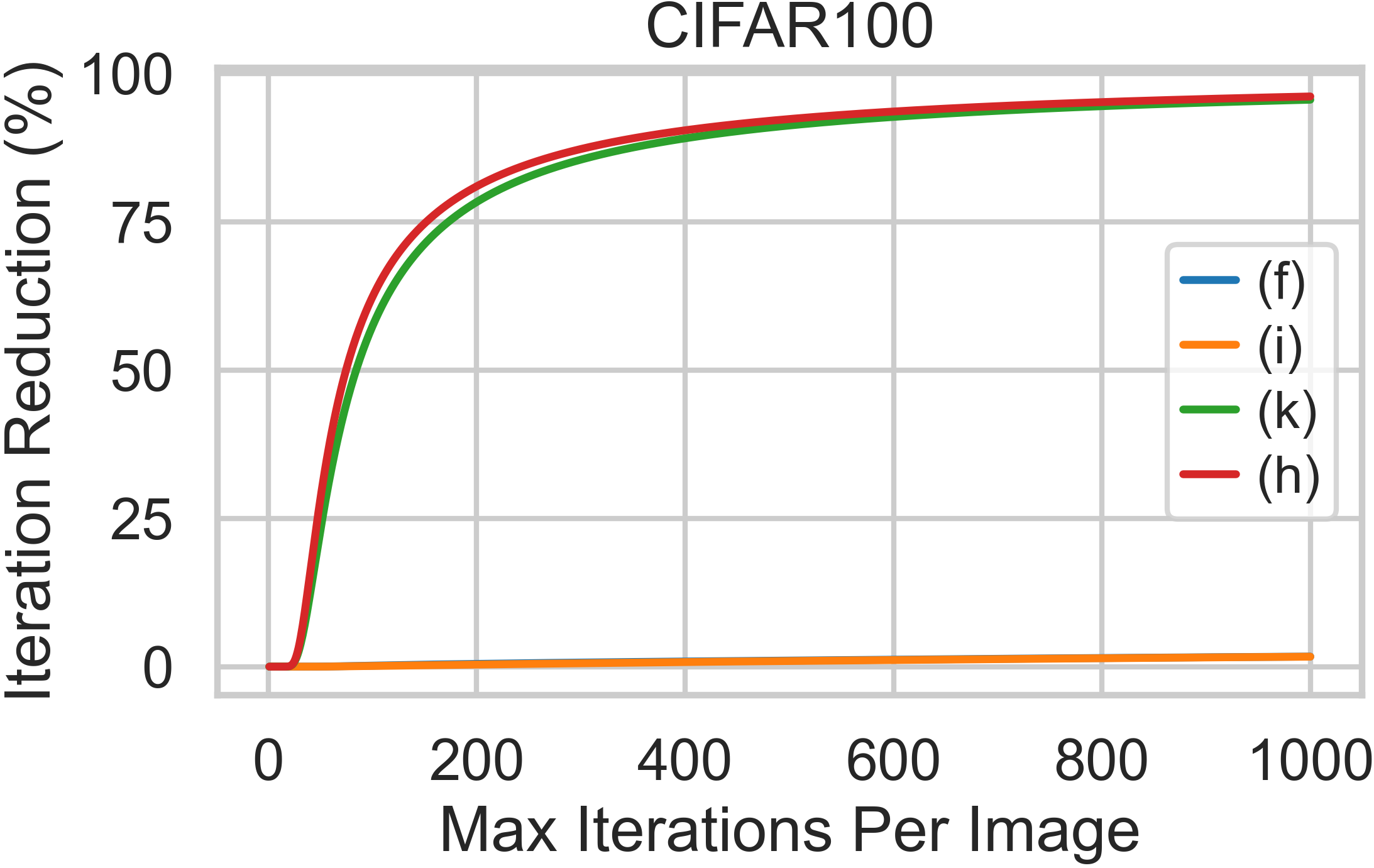}
  }
\\
\adjustbox{max width=\columnwidth}{%
  \includegraphics[]{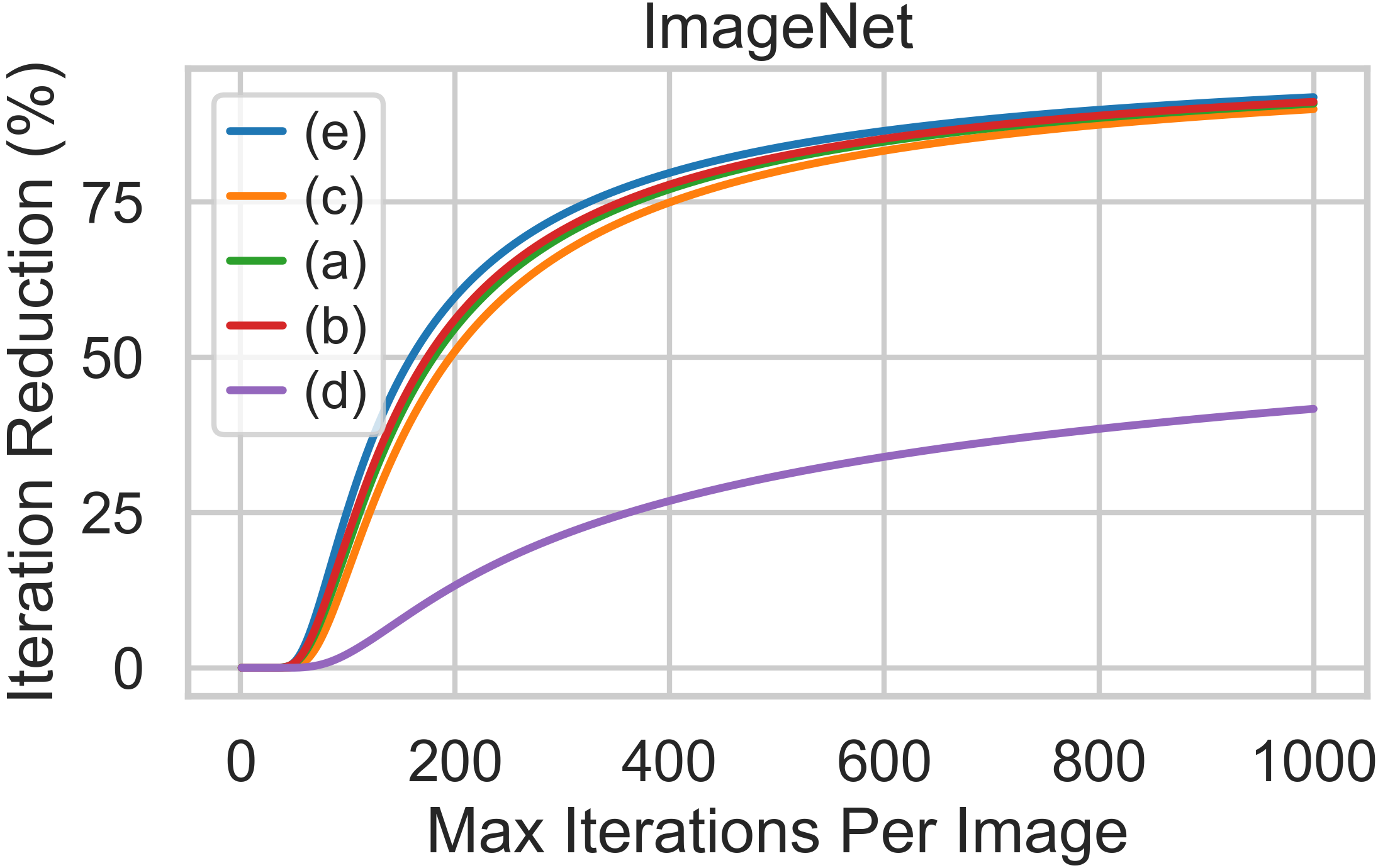}
  }
\caption{Plots of the percentage of iterations that cycle detection reduces in PGD against the maximum iteration budget per image, \(T_{\rm iter}\), for CIFAR10, CIFAR100, and ImageNet. Cycle detection obviously results in larger computational savings for larger iteration budgets, but most models enjoy significant computational savings even for reasonably small budgets like \(T_{\rm iter}=100.\)
}
\label{fig:reduction-vs-iteration-budget}
\end{figure}

In Table \ref{table:imagenet-cifar-saved-iterations} (the caption indicates which of 12 different models we tested) we record the clean accuracy of each model and the robust accuracy under the standard PGD attack as well as the robust accuracy under the PGD attack with cycle detection, denoted \(\rm PGD_{\rm CD}\). We display both values to emphasize that our cycle detection method gives the \emph{exact} same robust accuracy as standard PGD. More importantly, we also record the total number of PGD iterations required by standard PGD as well as by PGD with cycle detection to attack the entire dataset, 
and the computational savings by our cycle detection method. Additionally, in Figure \ref{fig:reduction-vs-iteration-budget} we plot the reduction of PGD iterations (as a percentage of the number of iterations used \emph{without} cycle detection) against \(T_{\rm iter}\) to demonstrate that our cycle detection technique often results in significant computational savings even for much smaller iteration budgets, for example \(T_{\rm iter} = 100\) achieves a 20-60\% reduction in several cases.

\begin{table}[!h]
\centering
\caption{
Results on ImageNet (rows a-e),
CIFAR100  (rows f-i), 
and CIFAR10 (rows j-l).
Accuracy for clean data, PGD, and our new  \(\rm PGD_{\rm CD}\) are shown. Note that  \(\rm PGD_{\rm CD}\) and PGD obtain the same accuracy, as explained in Section \ref{sec:cycle-detection}. The right three columns show that our cycle detection method obtains up to a 91\% reduction in the number of iterations required. 
The models, with their shortened names as presented/obtained from RobustBench \citet{croce2021robustbench}, are: 
(a) \emph{ConvNeXt-L} and (b) \emph{Swin-L} from \cite{liu2023comprehensive},
(c) \emph{XCiT-S12} \cite{debenedetti2023light}, 
(d) \emph{ViT-B-ConvStem} \cite{singh2024revisiting}, 
(e) \emph{MixedNUTS} \cite{bai2024mixednuts},
(f) \emph{Improving\_edm} \cite{bai2023improving}, 
(g) \emph{XCiT-L12} \cite{debenedetti2023light}, 
(h) \emph{WRN-70-16} \cite{wang2023better}, 
(i) \emph{Improving\_trades} \cite{bai2023improving},
(j) \emph{Carmon2019} \cite{carmon2019unlabeled}, 
(k) \emph{XCiT-L12} \cite{debenedetti2023light}, and 
(l) \emph{R18\_ddpm} \cite{rebuffi2021fixing}
.
}
\adjustbox{max width=\columnwidth}{%
\begin{tabular}{@{}lcccrrrr@{}}
\toprule
\multicolumn{1}{c}{}      & \multicolumn{3}{c}{Accuracy (\%)}    & \multicolumn{3}{c}{PGD Iterations}                                                                    & \multicolumn{1}{c}{Time}         \\ \cmidrule(lr){2-4} \cmidrule(lr){5-7} \cmidrule(l){8-8} 
\multicolumn{1}{c}{} & Clean & PGD   & \(\rm PGD_{\rm CD}\) & \multicolumn{1}{c}{PGD} & \multicolumn{1}{c}{\(\rm PGD_{\rm CD}\)} & \multicolumn{1}{c}{\%  $\downarrow$} & \multicolumn{1}{c}{\%  $\downarrow$} \\ \midrule
(a)                       & 77.48 & 58.15 & 58.15                & 29,110,933              & 2,681,126                                & 90.79                            & 90.53                            \\
(b)                       & 78.19 & 59.27 & 59.27                & 29,396,885              & 2,646,549                                & 91.00                            & 91.02                            \\
(c)                       & 72.54 & 42.70 & 42.70                & 21,409,926              & 2,160,645                                & 89.91                            & 89.44                            \\
(d)                       & 76.12 & 55.03 & 55.03                & 27,555,582              & 16,073,533                               & 41.67                            & 40.40                            \\
(e)                       & 81.11 & 67.52 & 67.52                & 33,784,085              & 2,758,488                                & 91.83                            & \textit{Error}                   \\ \midrule
(f)                       & 85.21 & 44.93 & 44.93                & 4,525,073               & 4,448,355                                & 1.70                             & -1.94                            \\
(g)                       & 70.76 & 38.93 & 38.93                & 3,903,289               & 170,518                                  & 95.63                            & 95.44                            \\
(h)                       & 75.22 & 48.19 & 48.19                & 4,826,965               & 184,368                                  & 96.18                            & 96.08                            \\
(i)                       & 80.18 & 40.22 & 40.22                & 4,052,162               & 3,984,016                                & 1.68                             & -0.28                            \\ \midrule
(j)                       & 89.69 & 62.01 & 62.01                & 6,214,308               & 3,371,294                                & 45.75                            & 35.40                            \\
(k)                       & 91.73 & 59.05 & 59.05                & 5,919,769               & 274,794                                  & 95.36                            & 94.97                            \\
(l)                       & 83.53 & 59.61 & 59.61                & 5,968,606               & 215,905                                  & 96.38                            & 93.15                            \\ \bottomrule
\end{tabular}
}
\label{table:imagenet-cifar-saved-iterations}
\end{table}

We see a significant reduction in PGD iterations in all of the ImageNet models we considered, in most cases reaching about a 90\% reduction, a $\approx10\times$ reduction. In the worst case, \(\rm PGD_{\rm CD}\) still achieves a 42\% reduction, saving a substantial amount of computational time at no downside. In the CIFAR10 models we considered, we see a similar story with some models attaining a reduction of \(>\)\(95\%\), resulting in a reduction by a factor of roughly 20-25$\times$ while still getting a reduction of about 45\% in the worst case. 

The CIFAR100 models represent our only ``failure'' case, in that we had no meaningful difference in computation time required for two models ($<2\%$) but substantial savings for the other two (\(>\)\(95\%\)). An important benefit of our approach is that these ``failures'' still obtain the same results, and take the same time to run, and so there is no negative consequence to using our method.

One key takeaway is obtained by observing the percentage reduction ($\% \downarrow$) in PGD iterations close tracks the time reduction using a some an unoptimized hash function\footnote{The \textit{Error} entry is due to a software conflict that occurred with a dependency update blocking model (e) from running at all. We could not resolve this in time for camera-ready. Timing numbers were measured on 1000 batches using 4 Tesla V100-SXM2-32GB GPU.}. At worst, $\rm PGD_{\rm CD}$ is the same speed as normal PGD, but in other cases, we save 10-20$\times$ on the total runtime. 
We emphasize, though, that these results are for robust inputs. For non-robust inputs (i.e., no cycle will form), our simple addition of a check for adversarial success also saves significant compute effort. 

In Table \ref{table:iterations-per-image} we record the mean and median number of iterations used by PGD with cycle detection on images that were tricked and untricked, as well as the overall statistics across all images. We see that the tricked images often only need a few iterations to trick, and this is the case both with and without cycle detection. We note that PGD without cycle detection will always use \(T_{\rm iter} = 1000\) iterations per untricked image and this is where cycle detection saves time. For tricked images, we can return after only 3 iterations for the plurality of images instead of 1000 onerous iterations. 

\begin{table}[!h]
\centering
\caption{Number of PGD iterations required per image on ImageNet (a-e), CIFAR100 (f-i), and CIFAR10 (j-l) with the same models from Table \ref{table:imagenet-cifar-saved-iterations}. We consider the mean and median number of iterations spent on each tricked image, each untricked image, and overall across all images. 
Images that are tricked can be detected with as few as 3 PGD iterations for all robust models, making a simple check save significant computational cost. Because many inputs are still tricked successfully, the overall average number of PGD iterations required is 1-2 orders of magnitude lower than what popular libraries perform.
}
\adjustbox{max width=\columnwidth}{%
\begin{tabular}{@{}lcccrrc@{}}
\toprule
\multicolumn{1}{c}{} & \multicolumn{2}{c}{Tricked} & \multicolumn{2}{c}{Untricked} & \multicolumn{2}{c}{Overall}\\ \cmidrule(lr){2-3} \cmidrule(lr){4-5} \cmidrule(lr){6-7}
\multicolumn{1}{c}{Model} & Mean & Median & Mean & Median & Mean & Median\\ \midrule
(a) & 3.61 & 3 & 91.01 & 82 & 69.21 & 71 \\
(b) & 3.58 & 3 & 88.16 & 79 & 67.70 & 68 \\
(c) & 3.88 & 3 & 98.48 & 88 & 59.57 & 64 \\
(d) & 3.66 & 3 & 582.73 & 578 & 422.34 & 181 \\
(e) & 3.84 & 3 & 80.94 & 72 & 68.02 & 67 \\
\cmidrule{1-7}
(f) & 7.96 & 2 & 982.92 & 1000 & 522.05 & 1000 \\
(g) & 3.23 & 3 & 41.16 & 37 & 24.10 & 25 \\
(h) & 2.95 & 3 & 36.61 & 34 & 24.51 & 28 \\
(i) & 7.54 & 3 & 983.06 & 1000 & 496.88 & 207 \\
\cmidrule{1-7}
(j) & 4.77 & 3 & 541.54 & 442 & 375.88 & 187 \\
(k) & 4.52 & 3 & 44.03 & 40 & 29.96 & 31 \\
(l) & 3.18 & 3 & 34.94 & 33 & 25.85 & 29 \\
\bottomrule
\end{tabular}%
} %
\label{table:iterations-per-image}
\end{table}

Although attack strength is not a primary concern in this work, as we instead focus on making the existing PGD attack more computationally efficient, we still note that all but one of our reported robust accuracies for ImageNet are within 1\% of those obtained by RobustBench which employs a more computationally demanding ensemble of attacks. Additionally, robust accuracies were all within 3\% of RobustBench for CIFAR10 and within 7\% of RobustBench for CIFAR100. These results imply that PGD with a fixed step size may provide tighter estimates of the robust accuracy of models that use higher-dimensional datasets, which further emphasizes the relevance of using PGD with cycle detection as a practical baseline attack on more realistic images. More importantly, PGD is still one of the most widely used attack strategies, which we can accelerate 
with our simple and easy-to-implement strategies.

\section{PGD can Still Beat Stronger Attacks}
\label{section:apgd}

Our $\text{PGD}_{\text{CD}}$ method is intrinsically useful in altering the Pareto-frontier of attack choices. Notionally, stronger attacks like Auto-PGD (APGD) that are more expensive but offer marginally higher attack success rates may be the inferior choice when considering compute time and cost. However, even though APGD uses methods like adaptive step sizes and momentum to improve upon PGD \cite{croce2020reliable}, we find that \textbf{PGD alone is often a more powerful attack than APGD} when we re-run these experiments over new architectures without parameter tunning. The speed efficacy of $\text{PGD}_{\text{CD}}$ is key in this consideration, as early termination allows the adaption and development of new algorithms that would be infeasible otherwise, and parameter running to improve attacks only increase the cost to perform them. As an example of this, we will further modify $\text{PGD}_{\text{CD}}$ to perform random restarts upon early termination, using a maximum of $T$ total iterations --- an attack strategy that is recommended but so expensive most papers do not perform such an evaluation. 

We demonstrate that $\rm PGD_{CD}$ often achieves a lower robust accuracy than APGD, and our cycle detection method also allows us to use significantly less computation.
Due to the aforementioned high computational cost, we will demonstrate our point using CIFAR10 and CIFAR100. Only comparing on CIFAR datasets is further justified by the fact that most of our \(\rm PGD_{\rm CD}\) results on ImageNet are already within 1\% of the reported RobustBench AutoAttack scores, so we know that the gap in results is, at best, small. 
By demonstrating that existing implementations of APGD, unmodified, have a lower attack success rate than $\text{PGD}_{\text{CD}}$, we show the practical utility and importance of PGD as a default benchmark method, and the need for its computational efficiency. This is not to say that APGD is always inferior in results, but that \textit{good and effective attack evaluation uses multiple tools. By making $\text{PGD}_{\text{CD}}$ both a strong baseline and the fastest tool, it should be the tool of first use --- as any image successfully attacked need not be considered with other slower alternatives like APGD.} 

Additionally, we note that \cite{croce2020reliable} originally compared APGD to PGD with a fixed step size by using 500 iterations with 5 random initializations (i.e. 100 iterations per run) per image. We want to make the following clear:

\begin{enumerate}
\item Cycle detection directly provides a \textbf{stronger} attack in the setting of PGD with random restarts and a fixed iteration budget if we wait to jump to a new random initialization once a cycle is detected. This means that the baseline that APGD was originally compared against in \cite{croce2020reliable} is a weaker baseline than what can be achieved with our cycle detection method, and thus the original comparison between APGD and PGD with a fixed step size was not a completely fair comparison.
\item Our cycle detection approach \emph{without} random restarts is still highly beneficial to use due to the tremendous computational savings it attains. Indeed, random restarts only offer marginal improvements to the reported robustness and APGD is often a \textbf{weaker} attack than PGD with a fixed step size, which we empirically demonstrate.
\end{enumerate}
In Table \ref{table:apgd-vs-pgd} we compare the reported robust accuracy from APGD with 500 iterations with 5 random restarts, PGD with random jumps after cycle detections (\(\rm PGD_{\rm CD}^{\rm jumps}\)) with \(T_{\rm iter}=500\), and PGD with cycle detection \emph{without} random jumps (\(\rm PGD_{\rm CD}\)) with \(T_{\rm iter}=500\). Additionally we report the number of iterations spent on images that were not misclassified (which makes up the vast majority of gradient computations, as misclassified images on average use very few iterations, as demonstrated in Table \ref{table:iterations-per-image}) to demonstrate \(\rm PGD_{\rm CD}\) uses significantly less compute in most cases while remaining competitive in attack strength. We used AutoAttack's implementation of APGD and, for simplicity, we only report APGD using the cross entropy loss as opposed to the DLR loss proposed in \cite{croce2020reliable}
as we empirically observed only marginal differences in reported robustness. For all CIFAR experiments we assume \(\epsilon=8/255\) and for PGD we use a fixed step size of \(\epsilon/4\). Indeed, we see that \(\rm PGD_{\rm CD}\) provides similar or better robust accuracies compared to other approaches while using substantially less compute.

\begin{table}[!h]
\centering
\caption{
Results on 
CIFAR100  (f-i), 
and CIFAR10 (j-l) with the same models from Table \ref{table:imagenet-cifar-saved-iterations}. Robust accuracy under APGD, \(\rm PGD_{\rm CD}^{\rm jumps}\), and \(\rm PGD_{\rm CD}\) are reported in addition to the number of iterations spent on images that were not misclassified by the attacks. While \(\rm PGD_{\rm CD}^{\rm jumps}\) offers marginal improvements in accuracy over \(\rm PGD_{\rm CD}\), we see that \(\rm PGD_{\rm CD}\) remains similar to or stronger than APGD while often being significantly less expensive.
}
\adjustbox{max width=\columnwidth}{%
\begin{tabular}{@{}lcccrrc@{}}
\toprule
\multicolumn{1}{c}{} & \multicolumn{3}{c}{Accuracy (\%)} & \multicolumn{3}{c}{PGD Iterations (Untricked)} \\ \cmidrule(lr){2-4} \cmidrule(lr){5-7}
\multicolumn{1}{c}{Model} & APGD & ${\rm PGD_{\rm CD}^{\rm jumps}}$ & ${\rm PGD_{\rm CD}}$ & \multicolumn{1}{c}{APGD} & \multicolumn{1}{c}{$\rm PGD_{\rm CD}^{\rm jumps}$} & ${\rm PGD_{\rm CD}}$ \\ \midrule
(f)  & 56.48 & 45.03 & 45.02 & 2,824,000 & 2,251,500 & 2,227,832\\
(g)  & 38.46 & 38.68 & 38.93 & 1,923,000 & 1,934,000 & 160,229\\
(h)  & 48.41 & 47.85 & 48.19 & 2,420,500 & 2,392,500 & 176,403\\
(i)  & 51.01 & 40.33 & 40.30 & 2,550,500 & 2,016,500 & 1,995,638\\
\cmidrule(lr){1-7}
(j)  & 69.35 & 61.92 & 62.02 & 3,467,500 & 3,096,000 & 2,140,630 \\
(k)  & 70.09 & 58.78 & 59.05 & 3,504,500 & 2,939,000 & 260,026\\
(l)  & 58.38 & 59.34 & 59.61 & 2,919,000 & 2,967,000 & 208,299 \\
\bottomrule
\end{tabular}%
}
\label{table:apgd-vs-pgd}
\end{table}

\section{Conclusion} 
\label{sec:conclusion}

In this work, we have presented a simple and novel method to significantly improve the computational feasibility of white-box adversarial robustness evaluation with PGD. 
Our cycle detection and termination method achieves $10 \mhyphen 20\times$ speedups in most cases 
while preserving the same attack strength as standard PGD. 
This work has a high impact on enabling the evaluation of adversarial robustness and comparisons of adversarial attacks at scale, 
by providing an attack method that produces identical success to ordinary PGD while providing significant computational efficiency.

Our improvements to PGD may enable other enhanced attack strategies as well. Our cycle detection method directly improves the attack strength (with respect to a fixed compute budget) of PGD with random initializations as one can choose to jump to a new random initialization as soon as a cycle is detected, allowing for more random initializations to be tested. Adversarial training could be accelerated similarly by stopping early, and any research or production need to evaluate a model will have reduced cost and carbon footprint.

\bibliography{references} %

\begin{thebibliography}{39}
\providecommand{\natexlab}[1]{#1}
\providecommand{\url}[1]{#1}
\csname url@samestyle\endcsname
\providecommand{\newblock}{\relax}
\providecommand{\bibinfo}[2]{#2}
\providecommand{\BIBentrySTDinterwordspacing}{\spaceskip=0pt\relax}
\providecommand{\BIBentryALTinterwordstretchfactor}{4}
\providecommand{\BIBentryALTinterwordspacing}{\spaceskip=\fontdimen2\font plus
\BIBentryALTinterwordstretchfactor\fontdimen3\font minus \fontdimen4\font\relax}
\providecommand{\BIBforeignlanguage}[2]{{%
\expandafter\ifx\csname l@#1\endcsname\relax
\typeout{** WARNING: IEEEtranN.bst: No hyphenation pattern has been}%
\typeout{** loaded for the language `#1'. Using the pattern for}%
\typeout{** the default language instead.}%
\else
\language=\csname l@#1\endcsname
\fi
#2}}
\providecommand{\BIBdecl}{\relax}
\BIBdecl

\bibitem[Szegedy et~al.(2013)Szegedy, Zaremba, Sutskever, Bruna, Erhan, Goodfellow, and Fergus]{szegedy2013intriguing}
C.~Szegedy, W.~Zaremba, I.~Sutskever, J.~Bruna, D.~Erhan, I.~Goodfellow, and R.~Fergus, ``Intriguing properties of neural networks,'' \emph{arXiv preprint arXiv:1312.6199}, 2013.

\bibitem[Zhang et~al.(2021)Zhang, Zheng, and Mao]{zhang2021adversarial}
X.~Zhang, X.~Zheng, and W.~Mao, ``Adversarial perturbation defense on deep neural networks,'' \emph{ACM Computing Surveys (CSUR)}, vol.~54, no.~8, pp. 1--36, 2021.

\bibitem[Liang et~al.(2022)Liang, He, Zhao, Jia, and Li]{liang2022adversarial}
H.~Liang, E.~He, Y.~Zhao, Z.~Jia, and H.~Li, ``Adversarial attack and defense: A survey,'' \emph{Electronics}, vol.~11, no.~8, p. 1283, 2022.

\bibitem[Akhtar et~al.(2021)Akhtar, Mian, Kardan, and Shah]{akhtar2021advances}
N.~Akhtar, A.~Mian, N.~Kardan, and M.~Shah, ``Advances in adversarial attacks and defenses in computer vision: A survey,'' \emph{IEEE Access}, vol.~9, pp. 155\,161--155\,196, 2021.

\bibitem[Costa et~al.(2024)Costa, Roxo, Proen{\c{c}}a, and In{\'a}cio]{costa2024deep}
J.~C. Costa, T.~Roxo, H.~Proen{\c{c}}a, and P.~R. In{\'a}cio, ``How deep learning sees the world: A survey on adversarial attacks \& defenses,'' \emph{IEEE Access}, 2024.

\bibitem[Chakraborty et~al.(2021)Chakraborty, Alam, Dey, Chattopadhyay, and Mukhopadhyay]{chakraborty2021survey}
A.~Chakraborty, M.~Alam, V.~Dey, A.~Chattopadhyay, and D.~Mukhopadhyay, ``A survey on adversarial attacks and defences,'' \emph{CAAI Transactions on Intelligence Technology}, vol.~6, no.~1, pp. 25--45, 2021.

\bibitem[Silva and Najafirad(2020)]{silva2020opportunities}
S.~H. Silva and P.~Najafirad, ``Opportunities and challenges in deep learning adversarial robustness: A survey,'' \emph{arXiv preprint arXiv:2007.00753}, 2020.

\bibitem[Athalye et~al.(2018)Athalye, Carlini, and Wagner]{athalye2018obfuscated}
A.~Athalye, N.~Carlini, and D.~Wagner, ``Obfuscated gradients give a false sense of security: Circumventing defenses to adversarial examples,'' in \emph{International conference on machine learning}.\hskip 1em plus 0.5em minus 0.4em\relax PMLR, 2018, pp. 274--283.

\bibitem[Carlini et~al.(2019)Carlini, Athalye, Papernot, Brendel, Rauber, Tsipras, Goodfellow, Madry, and Kurakin]{carlini2019evaluating}
N.~Carlini, A.~Athalye, N.~Papernot, W.~Brendel, J.~Rauber, D.~Tsipras, I.~Goodfellow, A.~Madry, and A.~Kurakin, ``On evaluating adversarial robustness,'' \emph{arXiv preprint arXiv:1902.06705}, 2019.

\bibitem[Pintor et~al.(2021)Pintor, Roli, Brendel, and Biggio]{NEURIPS2021_a709909b}
\BIBentryALTinterwordspacing
M.~Pintor, F.~Roli, W.~Brendel, and B.~Biggio, ``Fast minimum-norm adversarial attacks through adaptive norm constraints,'' in \emph{Advances in Neural Information Processing Systems}, M.~Ranzato, A.~Beygelzimer, Y.~Dauphin, P.~Liang, and J.~W. Vaughan, Eds., vol.~34.\hskip 1em plus 0.5em minus 0.4em\relax Curran Associates, Inc., 2021, pp. 20\,052--20\,062. [Online]. Available: \url{https://proceedings.neurips.cc/paper_files/paper/2021/file/a709909b1ea5c2bee24248203b1728a5-Paper.pdf}
\BIBentrySTDinterwordspacing

\bibitem[Mura et~al.(2025)Mura, Floris, Scionis, Piras, Pintor, Demontis, Giacinto, Biggio, and Roli]{MURA2025128918}
\BIBentryALTinterwordspacing
R.~Mura, G.~Floris, L.~Scionis, G.~Piras, M.~Pintor, A.~Demontis, G.~Giacinto, B.~Biggio, and F.~Roli, ``Ho-fmn: Hyperparameter optimization for fast minimum-norm attacks,'' \emph{Neurocomputing}, vol. 616, p. 128918, 2025. [Online]. Available: \url{https://www.sciencedirect.com/science/article/pii/S0925231224016898}
\BIBentrySTDinterwordspacing

\bibitem[Piras et~al.(2023)Piras, Floris, Mura, Scionis, Pintor, Biggio, and Demontis]{Piras2023}
\BIBentryALTinterwordspacing
G.~Piras, G.~Floris, R.~Mura, L.~Scionis, M.~Pintor, B.~Biggio, and A.~Demontis, ``Improving fast minimum-norm attacks with hyperparameter optimization,'' in \emph{ESANN 2023 proceesdings}, ser. ESANN 2023.\hskip 1em plus 0.5em minus 0.4em\relax Ciaco - i6doc.com, 2023. [Online]. Available: \url{http://dx.doi.org/10.14428/esann/2023.ES2023-164}
\BIBentrySTDinterwordspacing

\bibitem[Crecchi et~al.(2022)Crecchi, Melis, Sotgiu, Bacciu, and Biggio]{CRECCHI2022257}
\BIBentryALTinterwordspacing
F.~Crecchi, M.~Melis, A.~Sotgiu, D.~Bacciu, and B.~Biggio, ``Fader: Fast adversarial example rejection,'' \emph{Neurocomputing}, vol. 470, pp. 257--268, 2022. [Online]. Available: \url{https://www.sciencedirect.com/science/article/pii/S0925231221015708}
\BIBentrySTDinterwordspacing

\bibitem[Tjeng et~al.(2017)Tjeng, Xiao, and Tedrake]{tjeng2017evaluating}
V.~Tjeng, K.~Xiao, and R.~Tedrake, ``Evaluating robustness of neural networks with mixed integer programming,'' \emph{arXiv preprint arXiv:1711.07356}, 2017.

\bibitem[Rahnama et~al.(2020)Rahnama, Nguyen, and Raff]{Rahnama2020}
\BIBentryALTinterwordspacing
A.~Rahnama, A.~T. Nguyen, and E.~Raff, ``{Robust Design of Deep Neural Networks against Adversarial Attacks based on Lyapunov Theory},'' in \emph{The IEEE/CVF Conference on Computer Vision and Pattern Recognition (CVPR)}, 2020, pp. 8178--8187. [Online]. Available: \url{http://arxiv.org/abs/1911.04636}
\BIBentrySTDinterwordspacing

\bibitem[Cohen et~al.(2019)Cohen, Rosenfeld, and Kolter]{cohen2019certified}
J.~Cohen, E.~Rosenfeld, and Z.~Kolter, ``Certified adversarial robustness via randomized smoothing,'' in \emph{international conference on machine learning}.\hskip 1em plus 0.5em minus 0.4em\relax PMLR, 2019, pp. 1310--1320.

\bibitem[Salman et~al.(2019)Salman, Li, Razenshteyn, Zhang, Zhang, Bubeck, and Yang]{salman2019provably}
H.~Salman, J.~Li, I.~Razenshteyn, P.~Zhang, H.~Zhang, S.~Bubeck, and G.~Yang, ``Provably robust deep learning via adversarially trained smoothed classifiers,'' \emph{Advances in neural information processing systems}, vol.~32, 2019.

\bibitem[Raff et~al.(2019)Raff, Sylvester, Forsyth, and McLean]{Raff_BaRT_2019}
E.~Raff, J.~Sylvester, S.~Forsyth, and M.~McLean, ``{Barrage of Random Transforms for Adversarially Robust Defense},'' in \emph{The IEEE Conference on Computer Vision and Pattern Recognition (CVPR)}, Long Beach, CA, 2019, pp. 6528--6537.

\bibitem[Fleshman et~al.(2019)Fleshman, Raff, Sylvester, Forsyth, and McLean]{Fleshman2018a}
\BIBentryALTinterwordspacing
W.~Fleshman, E.~Raff, J.~Sylvester, S.~Forsyth, and M.~McLean, ``{Non-Negative Networks Against Adversarial Attacks},'' \emph{AAAI-2019 Workshop on Artificial Intelligence for Cyber Security}, 2019. [Online]. Available: \url{http://arxiv.org/abs/1806.06108}
\BIBentrySTDinterwordspacing

\bibitem[Lu et~al.(2022)Lu, Munoz, Fuchs, LeBlond, Zaresky-Williams, Raff, Ferraro, and Testa]{NEURIPS2022_1add3bbd}
\BIBentryALTinterwordspacing
F.~Lu, J.~Munoz, M.~Fuchs, T.~LeBlond, E.~Zaresky-Williams, E.~Raff, F.~Ferraro, and B.~Testa, ``A general framework for auditing differentially private machine learning,'' in \emph{Advances in Neural Information Processing Systems}, vol.~35.\hskip 1em plus 0.5em minus 0.4em\relax Curran Associates, Inc., 2022, pp. 4165--4176. [Online]. Available: \url{https://proceedings.neurips.cc/paper_files/paper/2022/file/1add3bbdbc20c403a383482a665eb5a4-Paper-Conference.pdf}
\BIBentrySTDinterwordspacing

\bibitem[Ganin et~al.(2016)Ganin, Ustinova, Ajakan, Germain, Larochelle, Laviolette, March, and Lempitsky]{ganin2016domain}
Y.~Ganin, E.~Ustinova, H.~Ajakan, P.~Germain, H.~Larochelle, F.~Laviolette, M.~March, and V.~Lempitsky, ``Domain-adversarial training of neural networks,'' \emph{Journal of machine learning research}, vol.~17, no.~59, pp. 1--35, 2016.

\bibitem[Goodfellow et~al.(2014)Goodfellow, Shlens, and Szegedy]{goodfellow2014explaining}
I.~J. Goodfellow, J.~Shlens, and C.~Szegedy, ``Explaining and harnessing adversarial examples,'' \emph{arXiv preprint arXiv:1412.6572}, 2014.

\bibitem[Carlini and Wagner(2017)]{7958570}
N.~Carlini and D.~Wagner, ``Towards evaluating the robustness of neural networks,'' in \emph{2017 IEEE Symposium on Security and Privacy (SP)}, 2017, pp. 39--57.

\bibitem[Madry et~al.(2017)Madry, Makelov, Schmidt, Tsipras, and Vladu]{madry2017towards}
A.~Madry, A.~Makelov, L.~Schmidt, D.~Tsipras, and A.~Vladu, ``Towards deep learning models resistant to adversarial attacks,'' \emph{arXiv preprint arXiv:1706.06083}, 2017.

\bibitem[Richards et~al.(2021)Richards, Nguyen, Capps, Forsythe, Matuszek, and Raff]{Richards2021}
\BIBentryALTinterwordspacing
L.~E. Richards, A.~Nguyen, R.~Capps, S.~Forsythe, C.~Matuszek, and E.~Raff, ``{Adversarial Transfer Attacks With Unknown Data and Class Overlap},'' in \emph{Proceedings of the 14th ACM Workshop on Artificial Intelligence and Security (AISec '21)}.\hskip 1em plus 0.5em minus 0.4em\relax Association for Computing Machinery, 2021. [Online]. Available: \url{http://arxiv.org/abs/2109.11125}
\BIBentrySTDinterwordspacing

\bibitem[Everett et~al.(2022)Everett, Nguyen, Richards, and Raff]{everett2022}
\BIBentryALTinterwordspacing
D.~Everett, A.~T. Nguyen, L.~E. Richards, and E.~Raff, ``Improving out-of-distribution detection via epistemic uncertainty adversarial training,'' 2022. [Online]. Available: \url{https://arxiv.org/abs/2209.03148}
\BIBentrySTDinterwordspacing

\bibitem[Chen et~al.(2022)Chen, Wang, Fan, Zhang, Li, and Lu]{Chen2022}
\BIBentryALTinterwordspacing
C.~Chen, Z.~Wang, Y.~Fan, X.~Zhang, D.~Li, and Q.~Lu, \emph{Nesterov Adam Iterative Fast Gradient Method for Adversarial Attacks}.\hskip 1em plus 0.5em minus 0.4em\relax Springer International Publishing, 2022, p. 586–598. [Online]. Available: \url{http://dx.doi.org/10.1007/978-3-031-15919-0_49}
\BIBentrySTDinterwordspacing

\bibitem[Kurakin et~al.(2018)Kurakin, Goodfellow, and Bengio]{kurakin2018adversarial}
A.~Kurakin, I.~J. Goodfellow, and S.~Bengio, ``Adversarial examples in the physical world,'' in \emph{Artificial intelligence safety and security}.\hskip 1em plus 0.5em minus 0.4em\relax Chapman and Hall/CRC, 2018, pp. 99--112.

\bibitem[Croce and Hein(2020)]{croce2020reliable}
F.~Croce and M.~Hein, ``Reliable evaluation of adversarial robustness with an ensemble of diverse parameter-free attacks,'' in \emph{International conference on machine learning}.\hskip 1em plus 0.5em minus 0.4em\relax PMLR, 2020, pp. 2206--2216.

\bibitem[Croce et~al.(2021)Croce, Andriushchenko, Sehwag, Debenedetti, Flammarion, Chiang, Mittal, and Hein]{croce2021robustbench}
\BIBentryALTinterwordspacing
F.~Croce, M.~Andriushchenko, V.~Sehwag, E.~Debenedetti, N.~Flammarion, M.~Chiang, P.~Mittal, and M.~Hein, ``{RobustBench: a standardized adversarial robustness benchmark},'' in \emph{Thirty-fifth Conference on Neural Information Processing Systems Datasets and Benchmarks Track}, 2021. [Online]. Available: \url{https://openreview.net/forum?id=SSKZPJCt7B}
\BIBentrySTDinterwordspacing

\bibitem[Moulay et~al.(2019)Moulay, L{\'e}chapp{\'e}, and Plestan]{moulay2019properties}
E.~Moulay, V.~L{\'e}chapp{\'e}, and F.~Plestan, ``Properties of the sign gradient descent algorithms,'' \emph{Information Sciences}, vol. 492, pp. 29--39, 2019.

\bibitem[Carmon et~al.(2019)Carmon, Raghunathan, Schmidt, Duchi, and Liang]{carmon2019unlabeled}
Y.~Carmon, A.~Raghunathan, L.~Schmidt, J.~C. Duchi, and P.~S. Liang, ``Unlabeled data improves adversarial robustness,'' \emph{Advances in neural information processing systems}, vol.~32, 2019.

\bibitem[Liu et~al.(2023)Liu, Dong, Xiang, Yang, Su, Zhu, Chen, He, Xue, and Zheng]{liu2023comprehensive}
C.~Liu, Y.~Dong, W.~Xiang, X.~Yang, H.~Su, J.~Zhu, Y.~Chen, Y.~He, H.~Xue, and S.~Zheng, ``A comprehensive study on robustness of image classification models: Benchmarking and rethinking,'' \emph{arXiv preprint arXiv:2302.14301}, 2023.

\bibitem[Debenedetti et~al.(2023)Debenedetti, Sehwag, and Mittal]{debenedetti2023light}
E.~Debenedetti, V.~Sehwag, and P.~Mittal, ``A light recipe to train robust vision transformers,'' in \emph{2023 IEEE Conference on Secure and Trustworthy Machine Learning (SaTML)}.\hskip 1em plus 0.5em minus 0.4em\relax IEEE, 2023, pp. 225--253.

\bibitem[Singh et~al.(2024)Singh, Croce, and Hein]{singh2024revisiting}
N.~D. Singh, F.~Croce, and M.~Hein, ``Revisiting adversarial training for imagenet: Architectures, training and generalization across threat models,'' \emph{Advances in Neural Information Processing Systems}, vol.~36, 2024.

\bibitem[Bai et~al.(2024)Bai, Zhou, Patel, and Sojoudi]{bai2024mixednuts}
Y.~Bai, M.~Zhou, V.~M. Patel, and S.~Sojoudi, ``Mixednuts: Training-free accuracy-robustness balance via nonlinearly mixed classifiers,'' \emph{arXiv preprint arXiv:2402.02263}, 2024.

\bibitem[Bai et~al.(2023)Bai, Anderson, Kim, and Sojoudi]{bai2023improving}
Y.~Bai, B.~G. Anderson, A.~Kim, and S.~Sojoudi, ``Improving the accuracy-robustness trade-off of classifiers via adaptive smoothing,'' \emph{arXiv preprint arXiv:2301.12554}, 2023.

\bibitem[Wang et~al.(2023)Wang, Pang, Du, Lin, Liu, and Yan]{wang2023better}
Z.~Wang, T.~Pang, C.~Du, M.~Lin, W.~Liu, and S.~Yan, ``Better diffusion models further improve adversarial training,'' in \emph{International Conference on Machine Learning}.\hskip 1em plus 0.5em minus 0.4em\relax PMLR, 2023, pp. 36\,246--36\,263.

\bibitem[Rebuffi et~al.(2021)Rebuffi, Gowal, Calian, Stimberg, Wiles, and Mann]{rebuffi2021fixing}
S.-A. Rebuffi, S.~Gowal, D.~A. Calian, F.~Stimberg, O.~Wiles, and T.~Mann, ``Fixing data augmentation to improve adversarial robustness,'' \emph{arXiv preprint arXiv:2103.01946}, 2021.

\end{thebibliography}

\end{document}